\documentclass[letterpaper, 10pt, conference]{ieeeconf}

\IEEEoverridecommandlockouts                              

\usepackage{graphics} 
\usepackage{graphicx,subfigure}
\graphicspath{{./figures/}}

\usepackage[noadjust]{cite}

\usepackage{hyperref}
\usepackage{amsmath} 
\usepackage{amssymb}  
\usepackage[algo2e]{algorithm2e}
\usepackage{algorithm}
\usepackage{algorithmic}
\usepackage{balance}

\title{\LARGE \bf
LiDAR and Inertial Fusion for Pose Estimation by Non-linear Optimization
}

\author{Haoyang Ye$^{1}$ and  Ming Liu$^{1}$
\thanks{*This paper is supported by the Research Grant Council of Hong Kong SAR Government, China, under project No. 16206014 and No. 16212815; National Natural Science Foundation of China No. 6140021318, awarded to Prof. Ming Liu}
\thanks{$^{1}$ Department of Electronic and Computer Engineering, The Hong Kong University of Science and Technology
{\tt\small hy.ye@connect.ust.hk}, {\tt\small eelium@ust.hk}}%
}

\begin{document}

\maketitle
\thispagestyle{empty}
\pagestyle{empty}

\begin{abstract}

Pose estimation purely based on 3D point-cloud could suffer from degradation, e.g. scan blocks or scans in repetitive environments. To deal with this problem, we propose an approach for fusing 3D spinning LiDAR and IMU to estimate the ego-motion of the sensor body. The main idea of our work is to optimize the poses and states of two kinds of sensors with non-linear optimization methods. On the one hand, a bunch of IMU measurements are considered as a relative constraint using pre-integration and the state errors can be minimized with the help of laser pose estimation and non-linear optimization algorithms; on the other hand, the optimized IMU pose outputs can provide a better initial for the subsequent point-cloud matching. The method is evaluated under both simulation and real tests with comparison to the state-of-the-art. The results show that the proposed method can provide better pose estimation performance even in the degradation cases.

\end{abstract}

\section{Introduction}
\subsection{Motivation}
\begin{figure}[!ht]
  \centering
   \subfigure[Side View]{\includegraphics[width = 0.45\columnwidth]{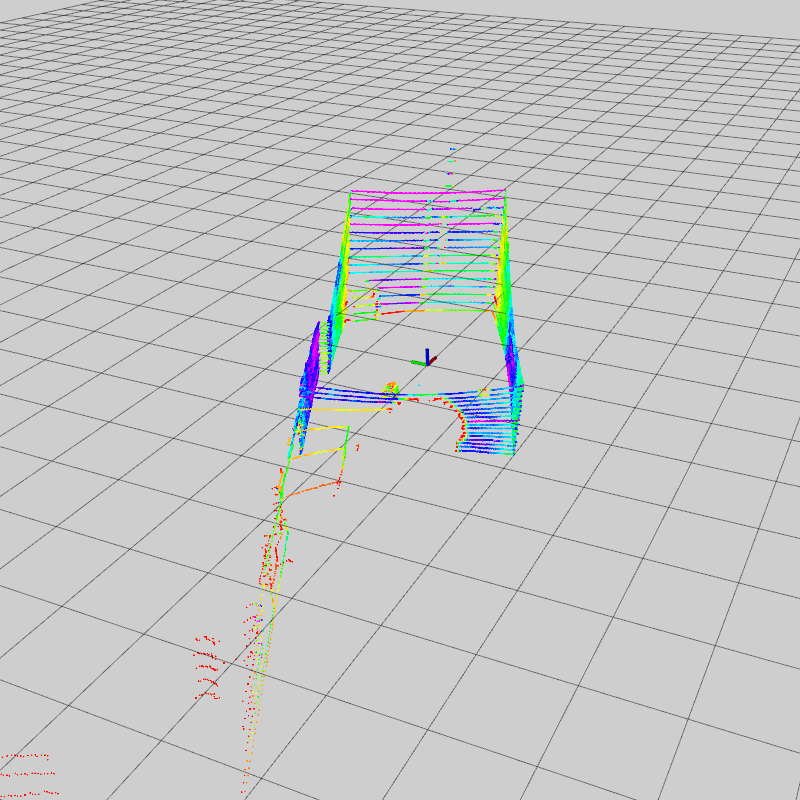}
    \label{fig:Side view}}
   \subfigure[Top View]{\includegraphics[width = 0.45\columnwidth]{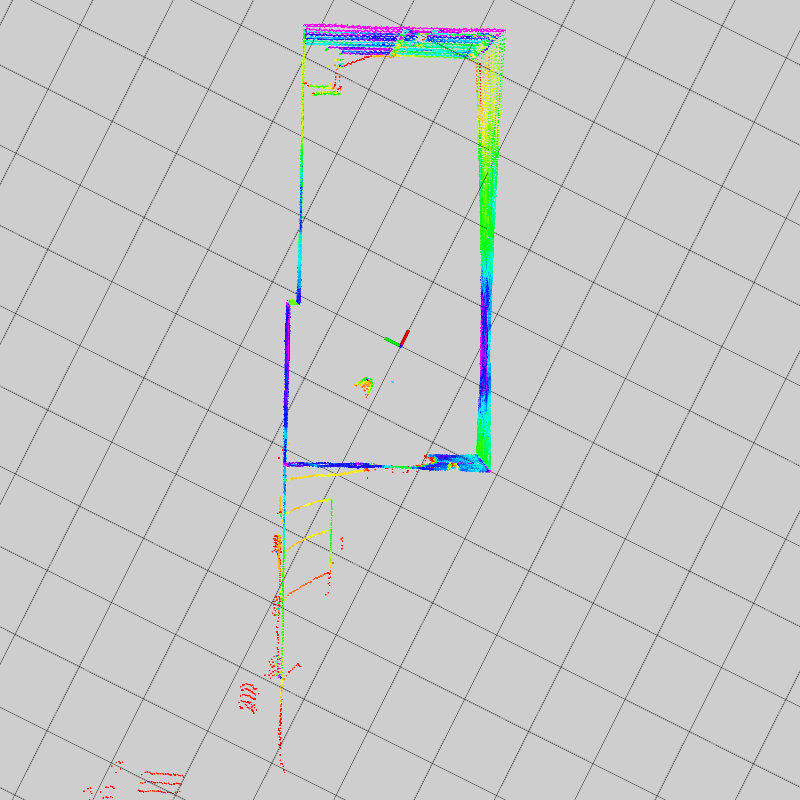}
    \label{fig:Top view}}
  \caption{Typical laser degradation case in corridor. There are few points observed on the ceiling or the ground, which could make the matching of point-clouds much challenging.}
    \label{narrow_corridor}
\end{figure}

Light Detection and Ranging (LiDAR) sensors can directly provide active high precision distance measurements to the surrounding objects. Due to the advantages of precision and the invariance to the environment illuminance, they have been widely applied in the fields of autonomous mobile robots. However, current algorithm to map sequencing scans may suffer from mismatches. Since the points from 3D spinning LiDAR scanner are typically sparse along its spinning axis, while dense in the plane perpendicular to the axis. For instance, a 16-beam 3D LiDAR has only 16 different rings vertically, but thousands of points per ring. Due to the directional limited points, the pose estimation of pitch angle and vertical axis could be problematic. In some situations, the 3D LiDAR can be partial blocked, which could cause the partial loss the point observations. For instance, in narrow corridors, lower parts of the point-cloud can disappear due to small response angles. Fig. \ref{narrow_corridor} depicts a typical case. In such cases, most of the points are around the sides, and few points on the ground or ceiling, which could bring matching errors to laser-only methods, such as iterative closest point (ICP)\cite{pomerleau2011tracking} and normal distributions transform (NDT)\cite{takeuchi20063}. At this time, the constraints would be insufficient for \textit{vertical translation} or \textit{pitch} rotation. Other cases such as shadow points or non-uniform density are not discussed here. Interested readers on these cases and mathematical solutions to these degradation are referred to \cite{liu2013icia}.

Another limitation of the pure point-cloud-based system is the update rate. For instance, most of the 3D spinning LiDARs can only provide 10-15 Hz updates, which are insufficient for the dynamic control problem. Inertial Measurement Unit (IMU) as an proprioceptive sensor is also invariant to light and other environments changes, which can provide relative fast output ($>$100 Hz). Thus, the need of combining 3D spinning LiDAR and IMU becomes intuitive.

Based on information fusion of the two sensors, the point-cloud matching error can be corrected aided by the collaboration of IMU. We develop a novel framework to obtain optimal pose estimation in this paper.

\subsection{Contribution}
The main contribution of our work is a framework that fuses point-cloud with IMU measurements. As output, the optimization-based pose estimation can be achieved. In our system, the pre-integration of IMU measurements \cite{forster2015imu} is adopted. Combined with the point-cloud-based pose estimation, the overall error is reduced by non-linear optimization. A more accurate initialization for point-cloud matching algorithms can be provided by the previously optimized IMU output poses. When matching error occurs in point-cloud matching estimation, the IMU-based estimation can comprise and then skip the bad temporal input of laser. For evaluation, we also compare our methods with non-fusion solutions, such as LOAM\cite{zhang2014loam} and our proposed method without the aid from IMU through simulation and experiments.


\subsection{Organization}
The rest of the paper is organized as follows: Section \ref{sec:related_work}  introduces related work about LiDAR pose estimation and sensor fusion; Model and processing about IMU and LiDAR are given in Section \ref{sec:imu_lidar_processing}; main framework of our method are stated in Section \ref{sec:framework_and_algorithms}; finally, we show the experiment results through simulation and real test in Section \ref{sec:experiments} and conlusion and future work in Section \ref{sec:conclusion_and_future_work}.

\section{Related Work}
\label{sec:related_work}
LiDAR as an accurate range sensor has been widely used in the filed of robotics, like action planning\cite{gianni2011unified}.  3D laser scans can be obtained from spinning 2D laser scanner with 3D scan-matching \cite{bosse2009continuous} or directly from 3D laser scanners. On 3D laser points, pose estimation can directly from laser, in \cite{pomerleau2014long}, ICP is used to do 3D point cloud matching. Works of fusion of 3D laser points and heterogenous sensors are also available. For example, Azim's method \cite{azim2012detection} applies IMU/GPS to directly get the transformation estimation of laser sensors. But his work does not use heterogenous sensors to obtain a better estimation of the poses directly from points.
\cite{balazadegan2016visual} proposed a visual laser odometry method aided by reduced IMU, where only 2-D acceleration (x and y axis) and 1-D gyro (z axis) information of IMU are used. It is shown that the fusion between 2D laser scanner and IMU can provide a better registration for the points in 3D \cite{zhang2014loam}. Then using the line and plane features from the obtained 3D laser points, robust pose estimation can be achieved.

One paradigm of fusion multiple sensors is the so-called loosely coupled extended Kalman filter (EKF) \cite{weiss2011real}. It takes the measurements from both sensors as blackbox, and use EKF to improve the quality of mapping and pose estimation. However, the initial states in this method would have a big influence on the results, and bad initial could cause the system failure. Another way of combining multiple sensor is proposed in Zhang et al.'s work \cite{zhang2017enabling}. It utilizes sequential pipeline to fuse vision, inertial and laser together. When some of the sensor failed, the system can bypass the failure and other sensors can still help to maintain a good pose estimation with aggressive motions.

This paper mainly discusses how to combine two types of sensors, LiDAR and IMU. Since both of them are light-invariant sensors, and they can work properly both day and night. Non-linear optimization method is applied to minimize the error in IMU states and the estimated poses. IMU can also provide fast pose outputs and better initial for point-cloud matching part, especially in narrow, structure-less and laser-blocked environments.

\section{IMU and LiDAR Measurement Processing}
\label{sec:imu_lidar_processing}

   \begin{figure}[thpb]
      \centering
      \includegraphics[width = 3in]{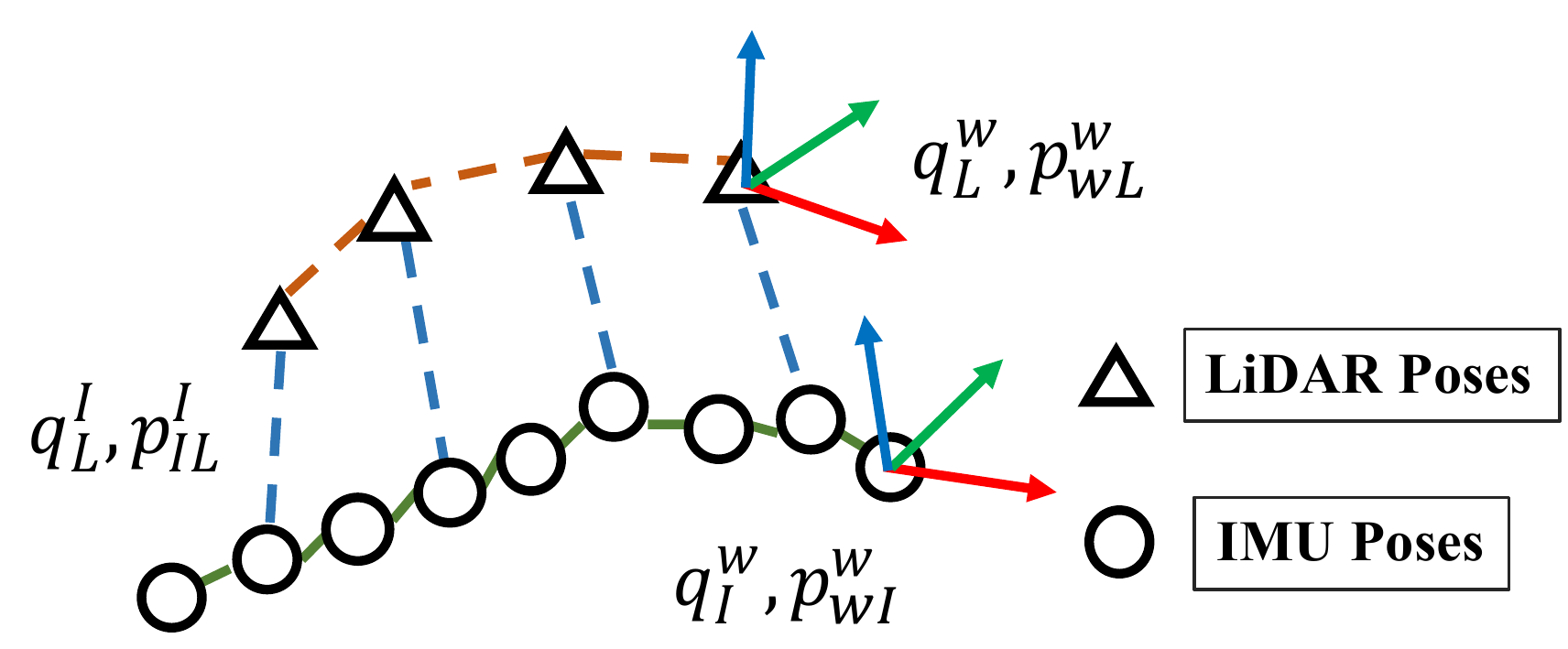}
      \caption{A diagram indicating the IMU and LiDAR pose sequences}
      \label{system_model}
   \end{figure}
In our system, IMU and laser are considered to be mounted on one rigid body, as a sensor body. The sensors will move and rotate with that rigid body. At the initial time, laser center is in the origin of the world frame. Fig. \ref{system_model} shows the setup of our system. There are three types of dash lines in the figure, which represent the translation and rotation between different frames. $q^w_{L}, p^w_{w_L}, q^w_{I}, p^w_{w I}, q^I_L, p^l_{IL} $ are rotations in quaternion and translations between two frames, from current laser frame to world frame, from current IMU frame to world frame, and from laser frame to IMU frame (extrinsic parameters) respectively.

\begin{figure*}[ht]
	\centering
	\includegraphics[width = 6in]{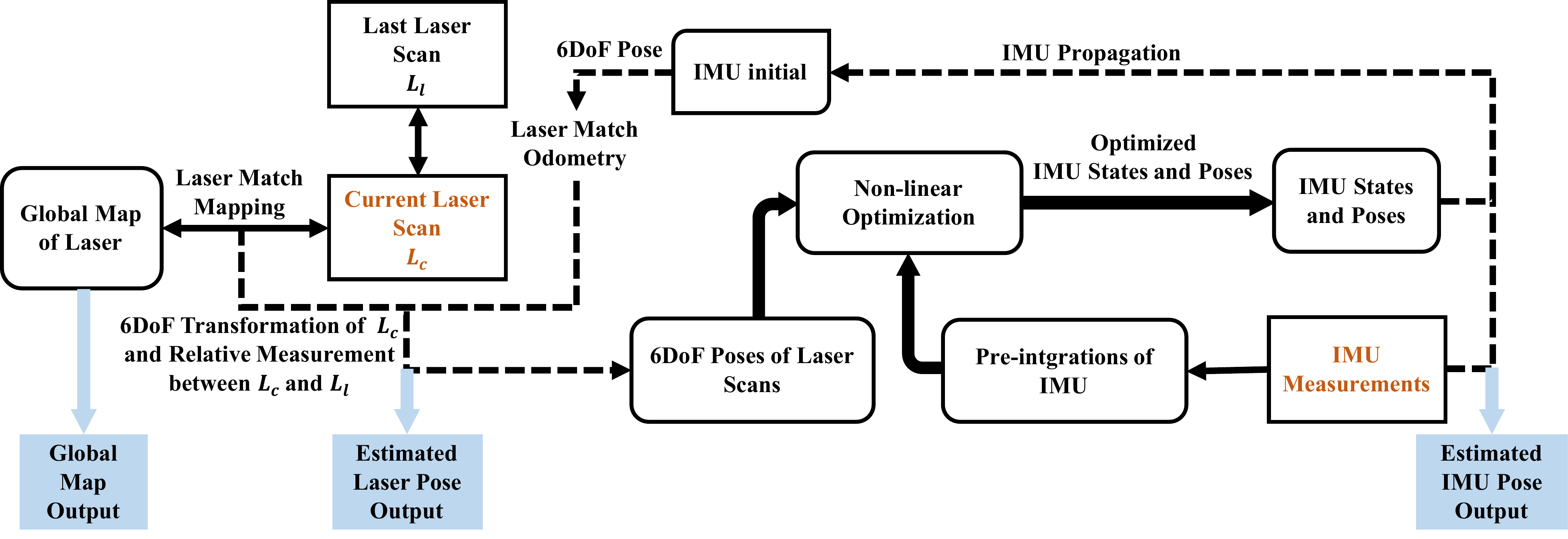}
	\caption{The overall framework of the proposed approach}
	\label{framework}
\end{figure*}

\subsection{IMU Model}
An simple IMU model is used in this paper, which is in accordance with \cite{weiss2011real}\cite{trawny2005indirect}, the real acceleration $a$ and angular velocity $\omega$ in current IMU frame are
\begin{equation} \label{eq_imu}
	a = a_m - b_a - n_a,\ \omega = \omega_m - b_\omega - n_\omega
\end{equation}
where the the measured values are denoted with subscript $m$, $b$ is the non-static bias, $n$ is the Gaussian noise in the measurement. Random processes are modeled for the non-static biases
\begin{equation} \label{eq_imu_bias}
	\dot{b_a} = n_{b_a},\ \dot{b_\omega} = n_{b_\omega}
\end{equation}

\subsection{IMU State Updates}
\label{subsec:imu_state_updates}
The IMU states we will estimate are the the rotations and translations in the Fig. \ref{system_model}, the velocity and biases, which are:
\begin{equation} \label{eq_X}
	X = [{p^w_{wI}}^T\ {v^w_{wI}}^T\ {q^w_I}^T\ {b_a}^T\ {b_\omega}^T]^T
\end{equation}
where ${p^w_{wI}}$, $v^w_{wI}$ are the position and velocity of IMU in world frame, $q^w_I$ is the rotation from current IMU frame to the world frame.

These states hold following differential equations:
\begin{align} \label{eq_de1}
	 \dot{p^w_{wI}} &= v^w_{wI} \\
	 \dot{v^w_{wI}} &= C_{(q^w_I)}(a_m-b_a-n_a)-g \\
        \dot{q^w_I} &= \frac{1}{2}\Omega_{JPL}(\omega_m-b_\omega-n_\omega) q^w_I \\
          \dot{b_a} &= n_{b_a},\ \dot{b_\omega} = n_{b_\omega}
\end{align}
where $C_{(q^w_I)}$ is the rotation matrix from quaternion, $\Omega_{JPL}$ is the left multiply matrix standing for the right multiply of quaternions, and $g$ is the gravity constant vector in world frame.

%

Then the states can be discretized following \cite{weiss2011real}\cite{trawny2005indirect}.
The estimated covariance matrix of states $P_{k+1}$ will be
\begin{equation} \label{eq_P_est}
	P_{k+1} = F_d P_{k} F_d^T + Q_d
\end{equation}
where $F_d$ is the discrete state update matrix of the states; $Q_d$ can be obtained using the continuous noise covariance matrix \cite{maybeck1982stochastic} $Q_c=\text{diag}(\sigma_{n_a}^2,\sigma_{n_{b_a}}^2,\sigma_{n_\omega}^2,\sigma_{n_{b_\omega}}^2)$, as 
\begin{equation} \label{eq_Qd}
	Q_d = \int_{\Delta t}F_d(\tau)G_cQ_C G_c^T F_d(\tau)^T d\tau
\end{equation}

\subsection{IMU Pre-integration}
Simply using IMU with discrete state propagation could cause repeated integration problem. Thus, we adopt IMU pre-integration from visual inertial methods \cite{forster2015imu, forster2017manifold, qin2017vins} into our approach. The key idea is to separate the the relative pose and velocity of IMU from the IMU states and regard a bunch of IMU measurements as one relative motion increment.
\begin{align}
\Delta p^{I_i}_{I_i I_j} &= R^T_{I_i}(p^w_{wI_j}-p^w_{wI_i}-v^w_{I_i}\Delta t_{ij} - \frac{1}{2}g\Delta t_{ij}^2) \\
\Delta v^{I_i}_{I_i I_j} &= R^T_{I_i}(v^w_{wI_j}-v^w_{wI_i}-g\Delta_{ij}) \\
\Delta R^{I_i}_{I_j} &= R^T_{I_i}R_{I_j}
\end{align}
where $i,j$ are different time stamp and $I_i,I_j$ denotes the frame of IMU at the corresponding time; $\Delta p^{I_i}_{I_i I_j}$, $\Delta v^{I_i}_{I_i I_j}$, $\Delta R^{I_i}_{I_j}$ are relative position, velocity and rotation increment respectively.
As described in \cite{forster2015imu, forster2017manifold, qin2017vins}, the above three states can be written as functions of acceleration bias $b_{a_{I}}$ and gyro biases $b_{g_{I}}$. Then the biases can also become an error term to be optimized, which will be discussed in Section \ref{sec:framework_and_algorithms}.

\subsection{LiDAR Measurements \& Poses}
Since the 3D spinning LiDAR could provide point-clouds as outputs. In this part, we will focus on laser-only matchings, which is a base of our method. There are plenty of point-cloud matching methods. Some methods extracts features from laser points to solve pose estimation problem. For instance, LOAM \cite{zhang2014loam} utilizes line and plane features, and \cite{dong2014lighting} suggests a reflection intensity based method. Some methods, like ICP, directly use the geometric points information to do the matching, instead of extracting features from point-clouds. In our method, point-to-plane ICP is used as our basic point-cloud matching strategy. And libpointmatcher \footnote{\url{https://github.com/ethz-asl/libpointmatcher}} is adopted to do ICP-based matching among different laser scans. We further modified it to be suitable for the calculations on sparse laser scans, which will be explained in detail in Section \ref{sec:framework_and_algorithms}.
After the matching, transformation matrix from the frame of current scan to the frame of last scan $T^{last}_{curr}$, and transformation matrix from the frame of current scan to the frame of global map $T^{last}_{global}$ can be obtained. $T^{last}_{curr}$ can be considered as relative measurement between two continuous frames of laser scans, and $T^{last}_{global}$ can provide the pose information of the current scan in the global frame.

\section{Framework and Algorithms}
\label{sec:framework_and_algorithms}

\subsection{Main Framework}

The framework of our method is shown in Fig. \ref{framework}. We can mainly separate our system into three parts, IMU part, LiDAR part, and optimization part.

In the IMU part, as shown in the right of Fig. \ref{framework}, the IMU states are first propagated from IMU measurements according to Section \ref{subsec:imu_state_updates}. The update rate in this part is much higher than LiDAR part, and can provide a initial for the point-cloud matching algorithm. At the meantime, bunches of IMU measurements are put into several pre-integration measurements, which can be used in the optimization part.

For LiDAR part, as shown in the left of Fig. \ref{framework}, the proposed LiDAR ICP algorithm pose estimation, in Section \ref{subsec:lidar_algorithm}, is based on IMU initial pose, current laser scan, measurement between current and last laser scan, and matching between current laser scan and global laser map. An estimated laser pose can be obtained from this part.

When both IMU and LiDAR are available, the optimization part (in the middle of Fig. \ref{framework}) will take effects. It will utilize 6DoF Poses of laser scans and IMU pre-integration data to correct IMU states, position $p^w_{wI_k}$, velocity $v^w_{wI_k}$, rotation $q^w_{I_k}$, and biases $b_{a_{I_k}}, b_{g_{I_k}}$. The subscript $k$ denotes different IMU pre-integration measurements. Then this part will output the optimized states of IMU, which are the corrected IMU outputs. Then these IMU outputs will be propagated with new IMU Measurements and a new cycle of sensor fusion will start.

\subsection{LiDAR Normal and Matching Algorithms}
\label{subsec:lidar_algorithm}
Point-to-plane ICP is chosen as our basic matching strategy. We then further improved it to be suitable for the point-clouds obtained be LiDAR. Because of the vertically sparse structure of LiDAR points, the typical normal calculation in point-to-plane ICP would be problematic. As shown in Fig. \ref{narrow_corridor}, the LiDAR points are in ring structures. The normal of one point can not just calculated based on K-nearest-neighbor points, since most of the nearest-neighbor points are prone to be in the same ring. Thus, the normal calculation from the whole point-cloud should be modified. We propose a LiDAR ICP matching algorithm, Algorithm \ref{alg:lidar normal}, which makes use of the ring information of scans from LiDAR sensor. The algorithm first select rings from the input points $L$. It can be obtained from LiDAR output or be calculated from the pitch angle of the points in $L$. Then the adjacent rings are used to find KNNs for each ring. Then, the KNNs can be used to calculate the surface normals of the points, by choosing the eigenvector with the smallest eigenvalue from the covariance matrix of the neighbor points \cite{RusuDoctoralDissertation}.

Algorithm \ref{alg:point-cloud matching} shows the LiDAR ICP Matching Algorithm, which does the matching between current laser scan $L_{curr}$ and laser laser scan $L_{last}$, as well as the matching between $L_{curr}$ and laser local map $L_{local}$. Here $L_{local}$ is cropped from the global map $L_{global}$ to reduce computational consumption. The inputs for the algorithm are initial transformation of $T_{init}$ from IMU propagation, transformation of last last scan to local frame $T^{local}_{last}$, $L_{curr}$, $L_{last}$, and $L_{local}$. The outputs are transformation from current scan frame to last scan frame $T^{last}_{curr}$, current scan frame to local frame $T^{local}_{curr}$ and the updated local map $L^{new}_{local}$. By two matching steps, shown as laser match odometry and laser match mapping in the left of Fig. \ref{framework}, our algorithm will calculate two transformations $T^{last}_{curr}$ and $T^{local}_{curr}$ sequentially. And if the error between $T^{local}_{curr}$ and its initial is too large, it is likely that mismatching in laser happens. The following merge and optimization parts will be skipped. Otherwise, if the matching works well, the algorithm will update the poses and map, and go into optimization part.

\begin{algorithm}
\KwData{$L$}
\KwResult{$L_{with\_normal}$}
\For{all points in current laser point-cloud $L$}
{
    Cast points into N different rings
}
\For{ring \textbf{i} in N rings}
{
    \textbf{j}, \textbf{k} = adjacent ring numbers of \textbf{i}\;
    
    Find \textbf{i}'s KNN in ring \textbf{j} and ring \textbf{k}, KNN\_j and KNN\_k\;
    
    Calculate point normal in ring \textbf{i}'s using KNN\_j and KNN\_k\;
    
    \textbf{Return} $L_{with\_normal}$ = $L$ with normals as point features
}
 \caption{LiDAR\_surface\_normal\_calculation}
 \label{alg:lidar normal}
\end{algorithm}

\begin{algorithm}
\KwData{$T_{init}$, $T^{local}_{last}$, $L_{curr}$, $L_{last}$, $L_{local}$}
\KwResult{$T^{last}_{curr}$, $T^{local}_{curr}$, $L^{new}_{local}$}

\If{System not initialized}
{
	Initialization
}

\textbf{Odometry:} $T^{last}_{curr}$ = Point-to-plane ICP for $L_{curr}$ and $L_{last}$ with $T_{init}$ as initial

\textbf{Mapping:} $T^{local}_{curr}$ = Point-to-plane ICP for $L_{curr}$ and $L_{local}$ with $T^{local}_{last}*T^{last}_{curr}$ as initial

\If{error between $T^{local}_{last}*T^{last}_{curr}$ and $T^{local}_{curr}$ $>$ threshold}
{
	Keep IMU poses unchanged and \textbf{return}
}
\Else
{
$L_{curr}$ = \textbf{LiDAR\_surface\_normal\_calculation}($L_{curr}$)\;

\textbf{Update} transformation $T^{last}_{curr}$, laser scan $L_{last}$\, \textbf{Transform} $L_{curr}$ with $T^{local}_{curr}$ into $L^{new}_{local}$

$\cdots$

\textbf{Into Optimization Part}
}

 \caption{LiDAR\_ICP\_matching Algorithm}
 \label{alg:point-cloud matching}
\end{algorithm}

\subsection{Non-linear Optimization Algorithm}

Non-linear Optimization Algorithm is used with IMU pre-integration and LiDAR poses. IMU 6DoF poses, velocities and biases are set as the parameters to be optimized. The constraints on the parameters are two kinds of measurements, IMU pre-integration measurements and relative LiRAR measurements.


In order to make the constraints related to both PIM and LiDAR measurements, we will optimize the IMU states at the time when we receive LiDAR data. The optimization constraints can be obtained as following. For the IMU measurement constraints, we adopt IMU pre-integration measurements (PIM) from visual inertial methods \cite{forster2015imu, forster2017manifold, qin2017vins} into our LiDAR-inertial framework.
The error of PIM is defined as
\begin{align}
States_i &= [p^w_{wI_i},v^w_{wI_i},R_{I_i},b_{a_{I_i}}, b_{g_{I_i}}] \\
States_j &= [p^w_{wI_j},v^w_{wI_j},R_{I_j},b_{a_{I_j}}, b_{g_{I_j}}] \\
PIM_{ij} &= [\Delta P^{I_i}_{I_i I_j}, \Delta v^{I_i}_{I_i I_j}, \Delta R^{I_i}_{I_j}] \\
 e_{PIM_{ij}} &= (States_i\oplus PIM_{ij}) - States_j
\end{align}
where $States_i$ and $States_j$ are the optimization parameters; $\oplus$ denotes the state update with PIM, the minus sign $-$ here can be regarded as $SE(3)$ inverse of composition operator for position and rotation, and subtraction for other states.
For the laser measurement constraints, a graph-based method \cite{grisetti2010tutorial} are used. Since LiDAR and IMU are fixed in the sensor body, in this paper the extrinsic parameters $q^I_L, p^l_{IL}$ are considered as fixed. Though we can still set them as optimization optional parameters. Based on the poses obtained from LiDAR, we can derive following formula to calculate the pose re-projection error, which is
\begin{align}
\hat{q^{L_i}_{L_j}} &= (q^w_{I_i}\otimes q^I_L)^{-1} \times (q^w_{I_j}\otimes q^I_L)\\
\hat{P^{w}_{L_iL_j}} &= P^{w}_{wL_i} + q^w_{I_i}P^I_L - (P^{w}_{wL_j} + q^w_{I_j}P^I_L)\\
\hat{P^{L_i}_{L_iL_j}} &= (q^w_{I_i}\otimes q^I_L)^{-1}\hat{P^{w}_{L_iL_j}}\\
 e_{Pose_{ij}} &=
 \begin{bmatrix}
 P^{L_i}_{L_iL_j} - \hat{P^{L_i}_{L_iL_j}}\\
 2.0\text{vec}(\hat{q^{L_i}_{L_j}}^{-1} \otimes q^{L_i}_{L_j}) 
 \end{bmatrix}
\end{align}

where the variables with hat sign are the estimated ones formed by re-projecting IMU poses into LiDAR measurements; the variables without hat sign are the measured values.
Thus, combinatingof those two errors together, we can get the final error among several continuous laser scans as

\begin{equation}
e_{total} = \sum\lVert e_{PIM_{ij}} \rVert_{2} + \sum\lVert e_{Pose_{ij}} \rVert_{2}
\end{equation}

Our goal is to optimize IMU 6DoF poses, velocities and biases to mininize $e_{total}$. For this non-linear optimization problem, Ceres Solver \footnote{\url{http://ceres-solver.org/}} is used to find the optimal solutions \cite{ceres-solver}.

\section{Experiments}
\label{sec:experiments} 
Both simulation and real experiments are implemented to validate our proposed system.
\subsection{Simulation Test}
\begin{figure}[thpb]
	\centering
	\includegraphics[width = 2.5in]{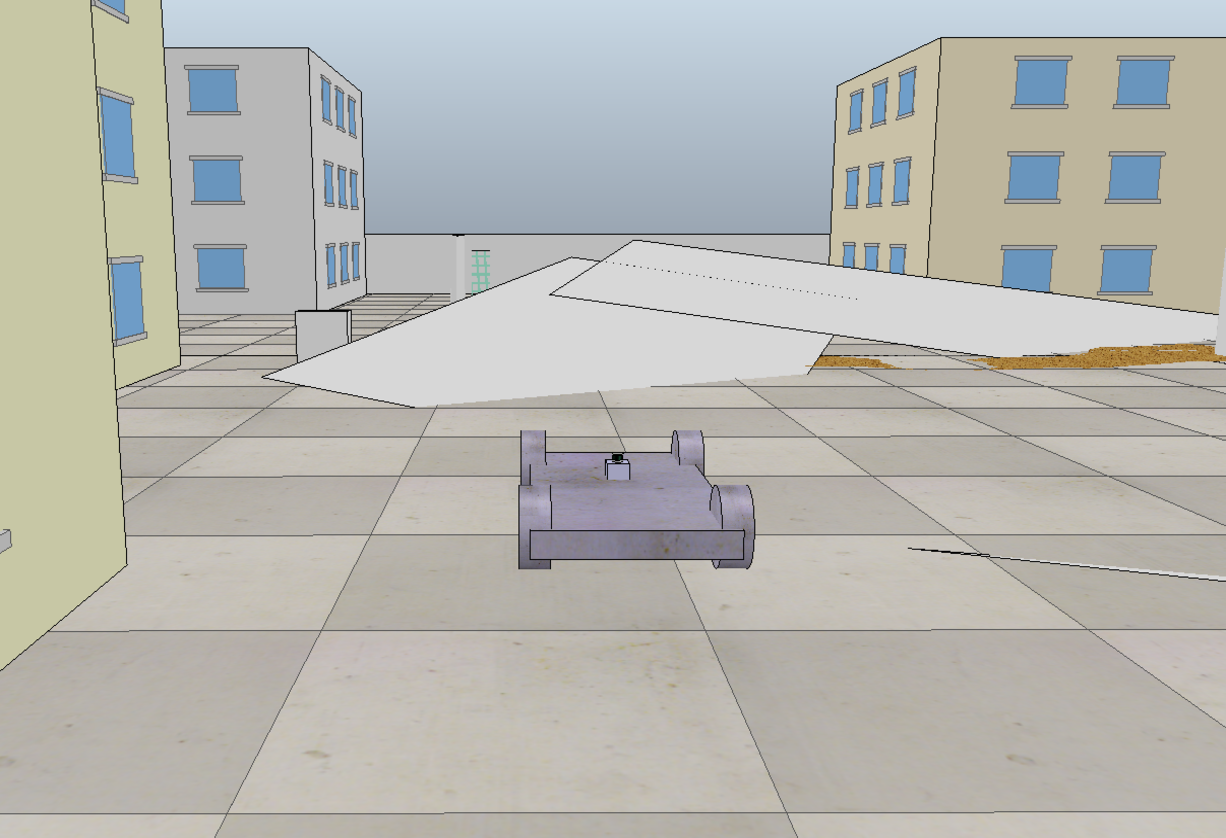}
	\caption{V-REP simulation. In the simulation, laser scanner and IMU are mounted on a moving vehicle to collect the sensor data and the ground truth.}
	\label{vrep_sim}
\end{figure}

Since it is difficult to get the ground truth of 6DoF motion, we first validate our method in simulation. We tried V-REP\footnote{\url{http://www.coppeliarobotics.com}} as our simulation platform. The sensor pair is mounted on a car in the simulation as shown in Fig. \ref{vrep_sim}. IMU data, laser scans and ground truth of sensor pair can be simulated in the platform.

To evaluate the estimation error, translation and rotation errors are separated \cite{geiger2012we}, which are defined as
\begin{eqnarray}
E_{rot}(\mathcal{F}) = \frac{1}{|\mathcal{F}|}\sum_{(i,j)\in\mathcal{F}}\angle[(\hat{p}_j\ominus\hat{p}_i)\ominus(p_j\ominus p_i)]\\
E_{trans}(\mathcal{F}) = \frac{1}{|\mathcal{F}|}\sum_{(i,j)\in\mathcal{F}}\lVert (\hat{p}_j\ominus\hat{p}_i)\ominus(p_j\ominus p_i)\rVert_{2}
\end{eqnarray}
where $i,j$ represent the frames in $\mathcal{F}$, $\hat{p} \in SE(3)$ is the estimated 6DoF pose of sensor body, and $p$ is the ground truth. The notation of $\ominus$ is the inverse of composition operator in $SE(3)$.

\begin{figure}[thpb]
	\centering
	\includegraphics[width = 3in]{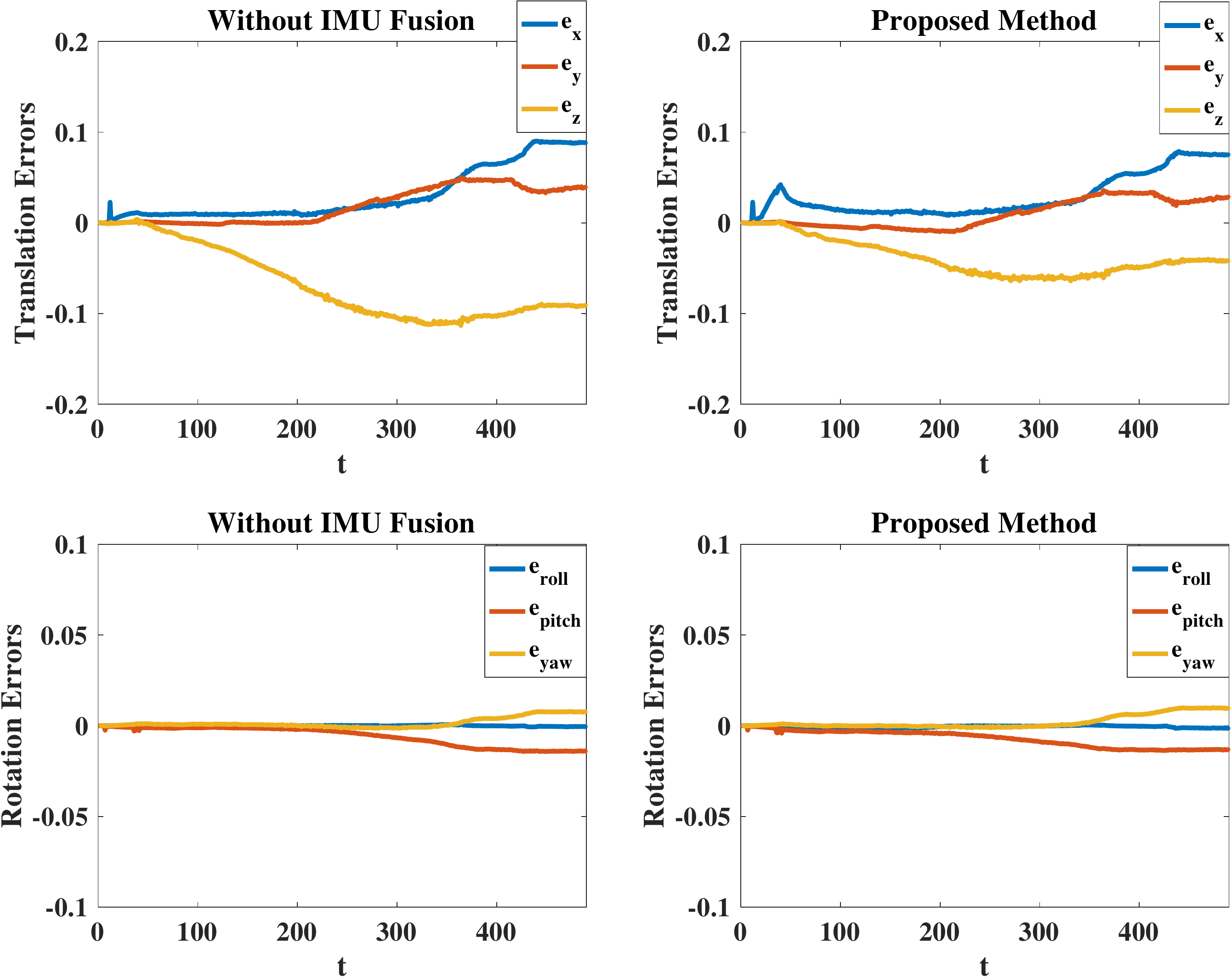}
	\caption{Translational and rotational errors along time. The top row shows the translation errors in x, y, and z axis; and the bottom row shows the rotation errors in roll, pitch and yaw. The left column depicts the results without fusing IMU data, while the right column shows the results of the proposed method.}
	\label{time_to_error}
\end{figure}

We average 245 frame pairs to compute the translation and rotation errors separately. The results are shown in Table \ref{table_sim_comparsion}. From the results we can find that rotation error in both methods are small, while for translation errors, the method with IMU fusion has a better performance. From Fig. \ref{time_to_error}, it shows that the method without IMU fusion could not have a poor performance especially on the z-axis translation. It also meets our expectation that for LiDAR data, laser-only method may not have good estimation on the vertical axis.

\begin{table}[!t]
\renewcommand{\arraystretch}{1.3}
\caption{Simulation Result Comparison}
\label{table_sim_comparsion}
\centering
\begin{tabular}{c||ccc}
\hline
\bfseries Method & \bfseries Without IMU Fusion & \bfseries Proposed Method \\
\hline\hline
Translation error & 3.0258 m & 1.9925 m \\
Rotation error & 0.3504$^{\circ}$ & 0.3574$^{\circ}$ \\
\hline
\end{tabular}
\end{table}

\subsection{Indoor \& Outdoor Experiments}
The proposed method was tested both indoor and outdoor on real data.  Both of the test datasets will be released on our lab website\footnote{\url{https://ram-lab.com/download/}}.

\subsubsection{Sensor Setup}
\begin{figure}[thpb]
	\centering
	\includegraphics[width = 2in]{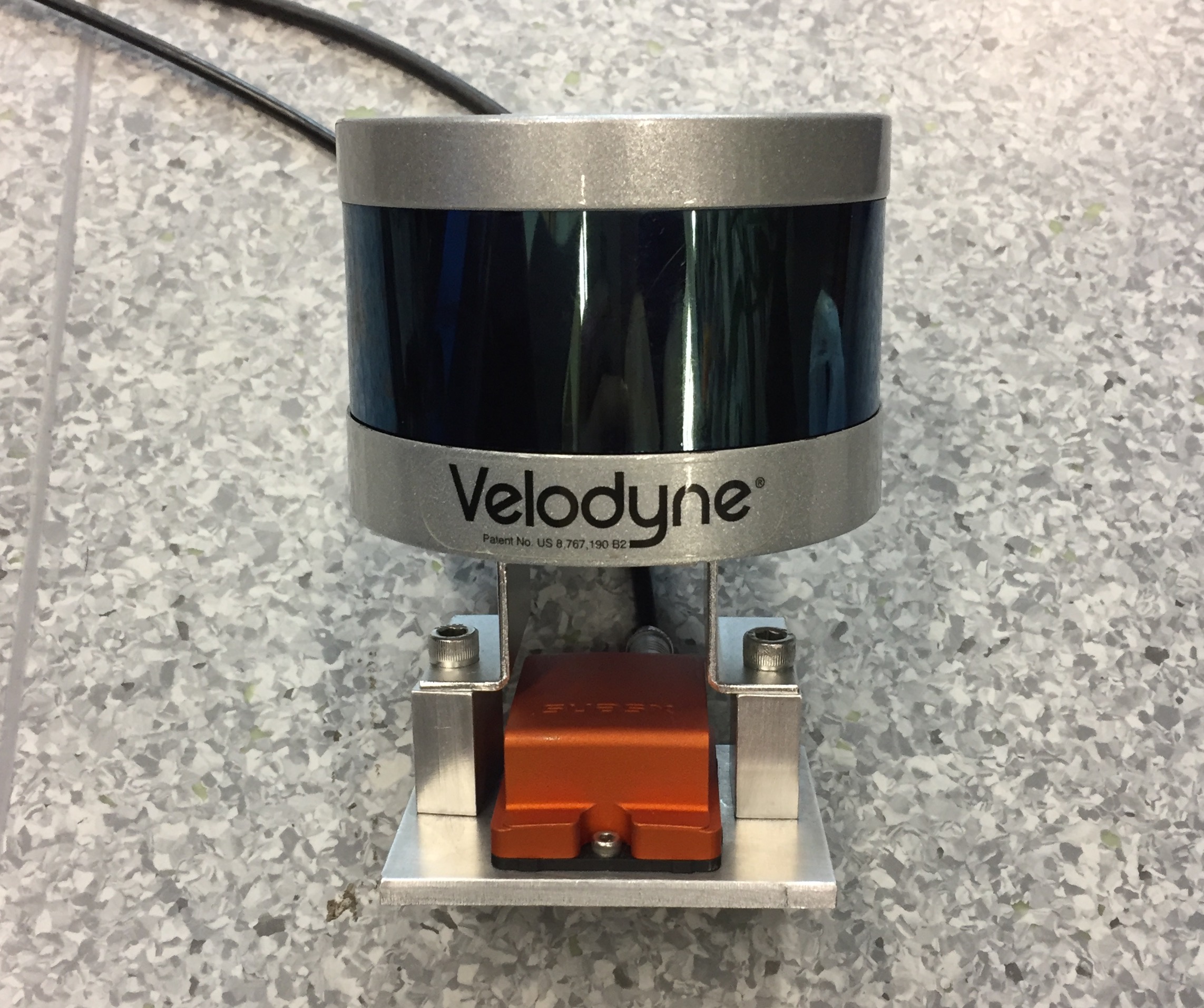}
	\caption{Sensor configuration. A 16-beam Velodyne VLP16 LiDAR sensor is mounted over a Xsens MTi-100 IMU sensor.}
	\label{sensor_setup}
\end{figure}

As shown in Fig. \ref{sensor_setup}, two sensors, a laser (Velodyne VLP-16\footnote{\url{http://velodynelidar.com/vlp-16.html}}) and a IMU (Xsens MTi 100\footnote{\url{https://www.xsens.com/products/mti-100-series/}}) are fixed on a rigid steel holder. Together, they can be regarded as a sensor body.

\subsubsection{Indoor Test}
For the indoor tests, the sensor body are hold by hand, point cloud and imu data were collected. Since the staircase and corridor are quite narrow, many of the scans from laser scanner might not contain points on horizontal planes. It makes the ego-motion estimation problem much harder.

\begin{figure}[thpb]
  \centering
  \includegraphics[width = 3in]{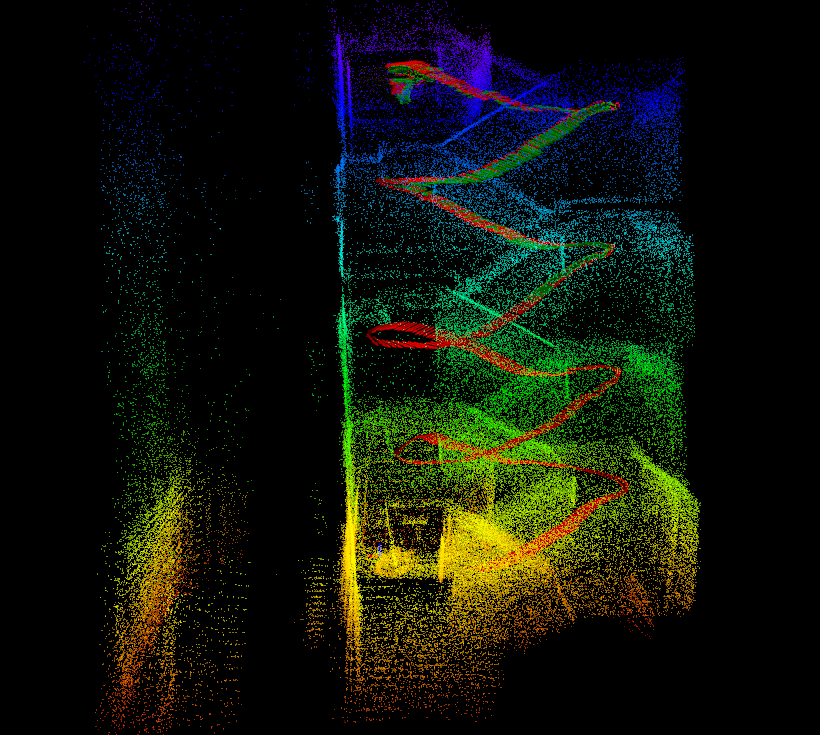}
  \caption{Indoor result of the proposed method. The red arrows indicate the trajectory of the sensor body, and the color points are the points of a staircase in 3D.}
  \label{fig:indoor_test}
\end{figure}

For comparison, our own datasets are tested by proposed methods as well as two non-IMU-aided methods, LOAM\footnote{\url{http://wiki.ros.org/loam_velodyne}}\cite{zhang2014loam} and our method without IMU fusion. As we can see from Fig. \ref{fig:LOAM, indoor}, such narrow case could cause failure in LOAM. Comparing the Fig. \ref{fig:Without IMU fusion, indoor} with Fig. \ref{fig:indoor_test}, it can be found that our proposed IMU-aided method can provide a more straight mapping of the staircase, which implies a better estimation on its ego-motion.

\begin{figure}[!ht]
  \centering
     \subfigure[LOAM]{\includegraphics[height = 0.4\columnwidth]{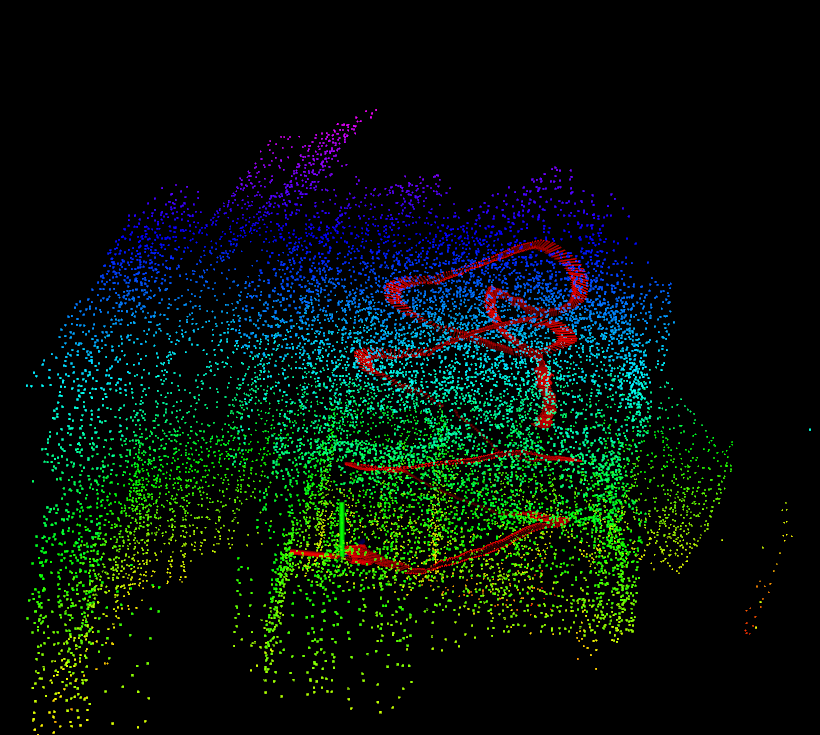}
    \label{fig:LOAM, indoor}}
      \subfigure[Without IMU fusion]{\includegraphics[height = 0.4\columnwidth]{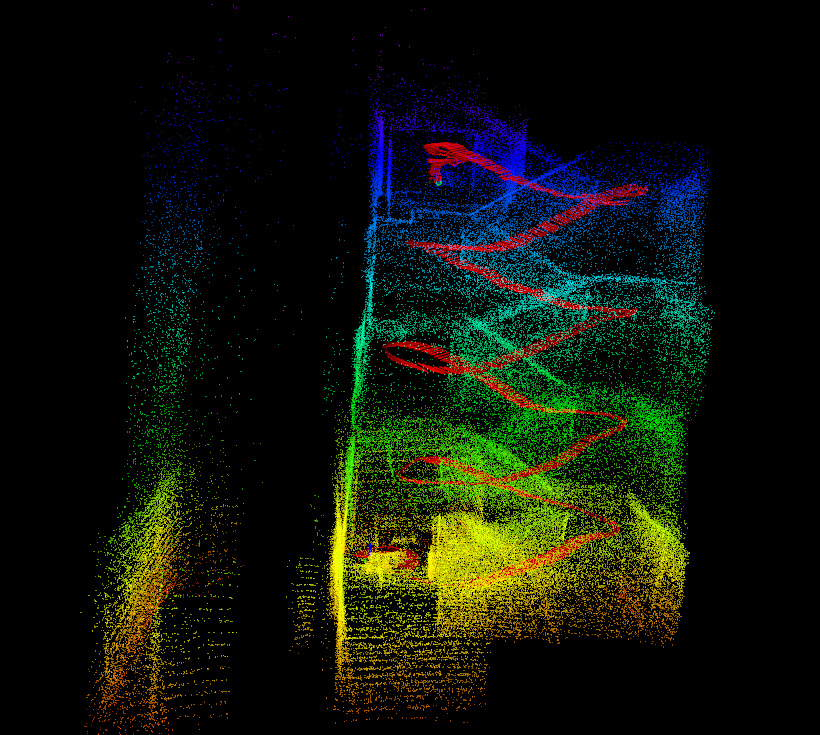}
    \label{fig:Without IMU fusion, indoor}}
  \caption{Comparative test results in the degraded cases. The left figure shows the results by LOAM and the right one shows the result of the proposed method without IMU fusion. The colorful points are the map built by two methods with the colormap in height. The red arrows indicate the corresponding odometry trajectories. Both figures are captured from the same point-of-view and scale.}
  \label{fig:indoor_comparoson}
\end{figure}

\subsubsection{Outdoor Test}
In the outdoor test, we tested our methods on the 16-beam laser with a low-cost IMU (MPU-9250\footnote{\url{https://www.invensense.com/products/motion-tracking/9-axis/mpu-9250/}}). At the meantime, laser points are half blocked, i.e. only the half front laser scans can be obtained. It will make the matching and pose estimation problems more challenge than the full scan cases.

\begin{figure}[!ht]
  \centering
     \subfigure[LOAM]{\includegraphics[height = 0.4\columnwidth]{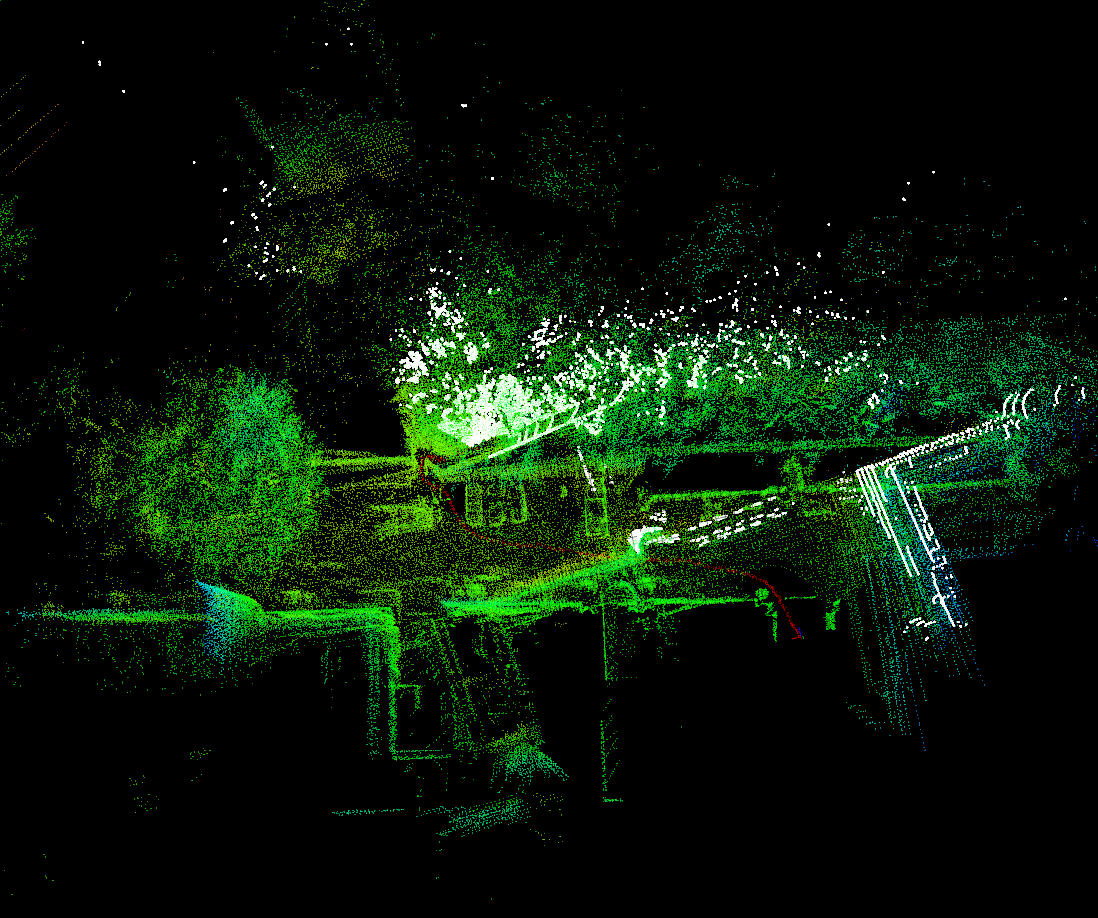}
    \label{fig:LOAM, outdoor}}
      \subfigure[Without IMU fusion]{\includegraphics[height = 0.4\columnwidth]{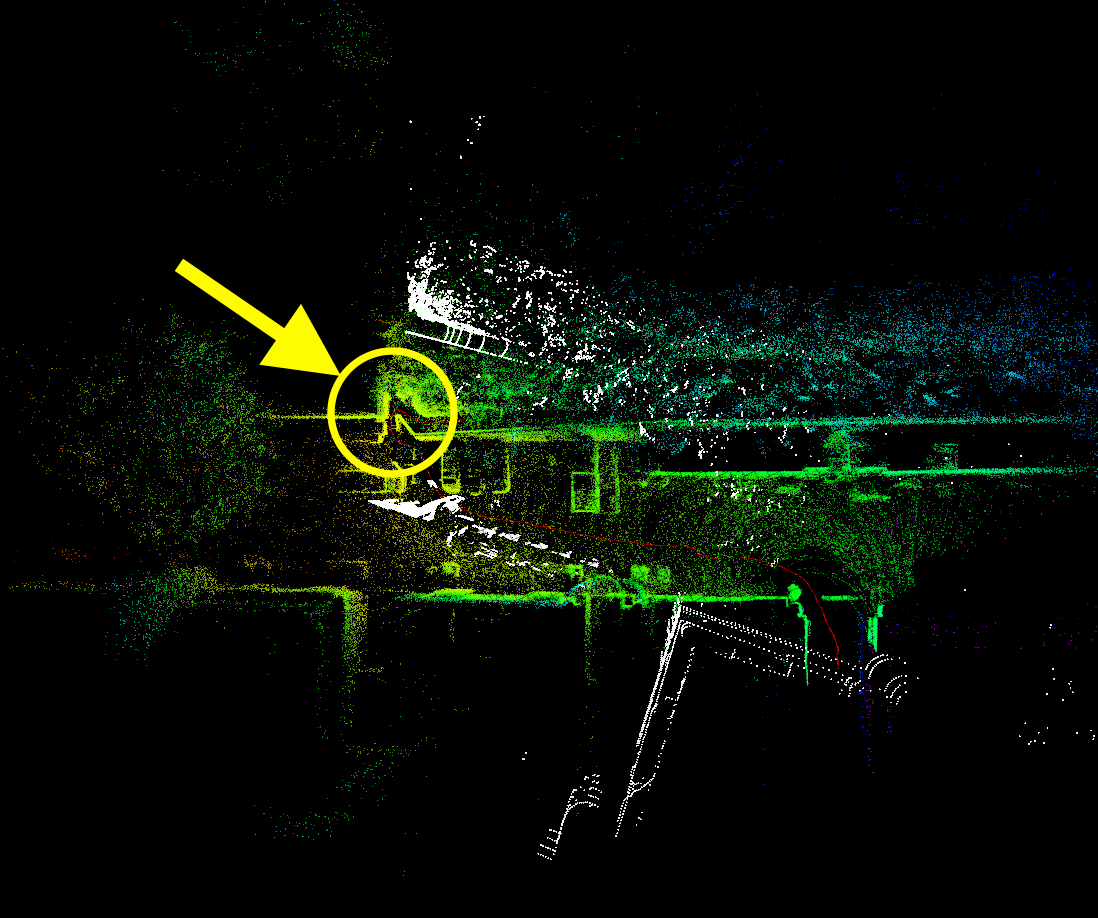}
    \label{fig:Without IMU fusion, outdoor}}
  \caption{Comparative results for an outdoor test. The left figure shows the results by LOAM and the right one shows the result of the proposed method without IMU fusion. Both results are not satisfied. The white points present the current laser scan in the 3D point cloud map.  Due to the narrow turn, as illustrated in yellow arrow and cycle, mismatches happen in both situations.}
  \label{outdoor_comparison}
\end{figure}

\begin{figure}[thpb]
  \centering
  \includegraphics[width = 3in]{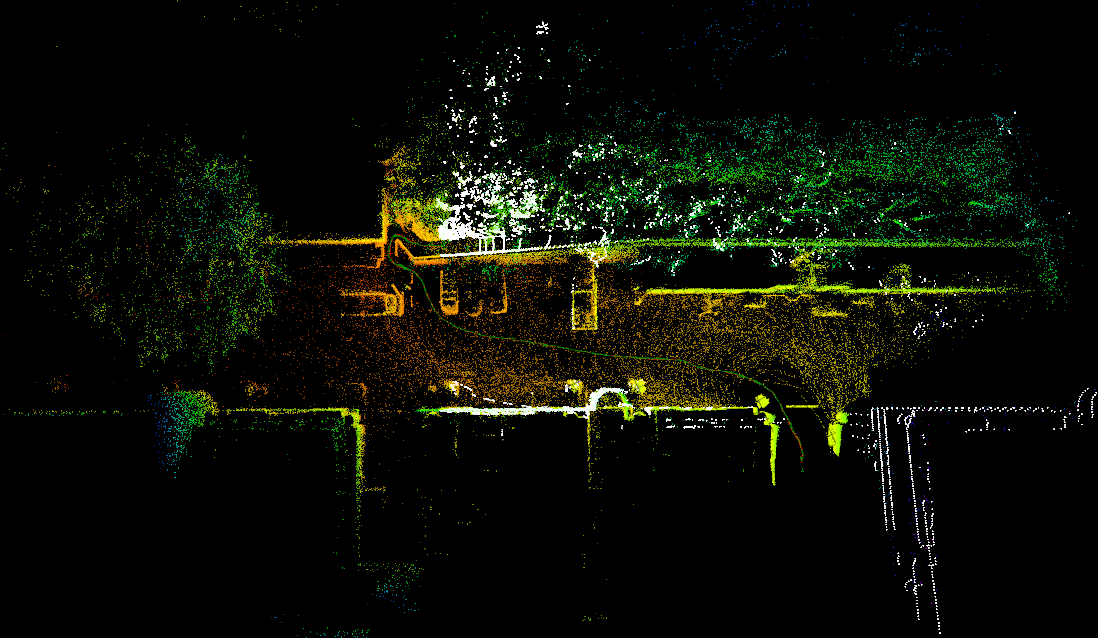}
  \caption{Outdoor test results. The green line is the IMU output pose trajectories, the colorful points are the points of outdoor environments in 3D and the white points are the current half front scan obtained from laser.}
  \label{outdoor_test}
\end{figure}
Fig. \ref{outdoor_comparison} and Fig. \ref{outdoor_test} show the results of outdoor tests. In this LiDAR degradation cases, our proposed method have a better performance than the other two methods. The non-IMU-aided methods suffer from too few constrains and points on a narrow turn, as highlighted in Fig. \ref{fig:Without IMU fusion, outdoor}. The point-cloud matching is likely to fail without any external aid, which causes the mismatching white points in Fig. \ref{outdoor_comparison}.

\section{Conclusion and Further Work}
\label{sec:conclusion_and_future_work}
Pose estimation from LiDAR sensor can be difficult when the laser points are degraded, because it can not provide enough useful points for the laser-only matching methods. Our proposed approach fused IMU and LiDAR to tackle this problem. In the system, the high-frequency updated IMU data could provide a good initial conditions for the point-cloud matching and overall map generation, even in the laser degradation cases; and relative robust poses could be obtained from the improved LiDAR matching method. Then an optimization-based algorithm would use the IMU and laser measurements from previous parts to maintain corrected IMU states. The method was tested both in simulation and real experiments and showed the advantages on degraded datasets.

Since pose graph method is used in our approach, we can further extend our work to enable the ability of loop closure detection and even merge more 3-D Maps \cite{bonanni20173}. The distortion from motion of the sensor is not considered in the paper, but it can also be implemented by registration points using fast pose estimation from IMU.









\bibliographystyle{IEEEtran}
\bibliography{IEEEabrv,optimal_laser_imu_bibfile}
\balance

\end{document}